\documentclass[runningheads]{llncs}

\usepackage{pdfpages}

\usepackage{booktabs}
\usepackage{graphicx}

\usepackage{float}
\usepackage{graphicx}
\usepackage{comment}
\usepackage{amsmath,amssymb} %
\usepackage{color}
\usepackage{xspace}
\usepackage{textcomp}
\usepackage{cite}
\usepackage{mathtools}
\usepackage{bbm}
\usepackage{makecell, verbatim, sidecap}
\usepackage{subfig}
\usepackage{multirow}
\usepackage{pifont}
\usepackage[capitalize]{cleveref}
\crefname{section}{Sec.}{Secs.}
\Crefname{section}{Section}{Sections}
\Crefname{table}{Table}{Tables}
\crefname{table}{Table}{Tabs.}

\def\ie{{i.e.\xspace}}

\def\cf{\emph{cf.}\xspace}

\def\handle{{Poseur}\xspace}
\def\R{{\mathbb R}}
\def\Loc{{\boldsymbol{\mu}}}
\def\loc{{\boldsymbol{\mu}}}
\def\a{\mathbf{a}}
\def\b{{\boldsymbol{b}}}
\def\KPF{{\hat{\mathbf{\Loc}}_f}}

\def\kpn{{\hat{\mathbf{\loc}}_n}}
\def\KBF{{\hat{\mathbf{\b}}_f}}
\def\KPQ{{\hat{\mathbf{\Loc}}_q}}

\def\KBQ{{\hat{\mathbf{\b}}_q}}
\def\Q{{\mathbf{Q}}}
\def\QZ{{\Q_z}}
\def\Qc{{\Q_c}}

\def\p{{\mathbf{p}}}
\def\x{{\mathbf{x}}}
\def\A{{\mathbf{A}}}
\def\W{{\mathbf{W}}}
\def\BTheta{{\boldsymbol{\Theta}}}
\def\BPhi{{\boldsymbol{\Phi}}}
\newcommand{\cmark}{\ding{51}}%
\newcommand{\xmark}{\ding{55}}%

\begin{document}
\pagestyle{headings}
\mainmatter
\def\ECCVSubNumber{4552}  %

\title{Poseur: Direct Human Pose Regression with Transformers\thanks{WM and YG contributed equally.
YG's contribution was in part made when visiting Alibaba.
ZT is now with Meituan Inc.
CS is the corresponding author.
\it 
Accepted to Proc.\ Eur. Conf. Computer Vision 2022.
}} %

\author{Weian Mao\textsuperscript{$1$}
        ~ Yongtao Ge\textsuperscript{$1$}
        ~ Chunhua Shen\textsuperscript{$3$} 
        ~ Zhi Tian\textsuperscript{$1$} 
        ~ Xinlong Wang\textsuperscript{$1$} 
        ~ Zhibin Wang\textsuperscript{$2$} 
        ~ Anton van den Hengel\textsuperscript{$1$} }
\institute{$^1$ The University of Adelaide
~
$^2$ Alibaba Damo Academy 
~
$^3$ Zhejiang University }

\authorrunning{W.\  Mao et al.}

%
%
\maketitle

\begin{abstract}
We propose a 
direct,
regression-based approach to 2D human pose estimation from single images.
We formulate the problem as a sequence prediction 
task,  
which we solve using a Transformer network. 
This network {\it directly} learns a regression mapping from images to the keypoint coordinates, without resorting to intermediate representations such as heatmaps.
This approach avoids much of the complexity associated with heatmap-based approaches. 
To overcome the feature misalignment issues 
of previous regression-based methods, 
we propose an attention mechanism that adaptively attends to the features
that are 
most relevant to the target keypoints, 
considerably improving the  accuracy.
Importantly, our framework is end-to-end differentiable, and naturally learns to exploit the %
dependencies 
between keypoints.
Experiments on MS-COCO and MPII, two predominant pose-estimation datasets, demonstrate that our method significantly improves upon the state-of-the-art in regression-based pose estimation.
More notably, ours is the first regression-based approach to perform
favorably 
compared to the best heatmap-based pose estimation methods. 
Code is available at:
\def\UrlFont{\tt \color{blue}}
{\url{\tt https://github.com/aim-uofa/Poseur}}

\keywords{2D Human Pose Estimation, Keypoint Detection, Transformer}

\end{abstract}

\section{Introduction}

Human pose estimation is one of the core challenges in computer vision, not least due to its importance in understanding human behaviour.  It is also a critical pre-process to a variety of human-centered challenges including activity recognition, video augmentation, and human-robot interaction.
Human pose estimation 
requires estimating the location of a set of keypoints in an image, in order that the pose of a simplified human skeleton might be recovered.

Existing methods for human pose estimation can be broadly categorized into heatmap-based and regression-based methods.
Heatmap-based methods first predict a heatmap, or classification score map, that reflects the likelihood that each pixel in a region corresponds to a particular skeleton keypoint. The current state-of-the-art methods use a fully convolutional network (FCN) to estimate this heatmap. The final keypoint location estimate corresponds to the peak in the heatmap intensity. Most current pose estimation methods are heatmap-based because this approach has thus far achieved higher accuracy than regression-based approaches. Heatmap-based methods have their disadvantages, however. 
1) The ground-truth heatmaps need to be manually designed and heuristically tuned.  The noise inevitably introduced impacts on the final results~\cite{luo2020rethinking,sun2017compositional,huang2020devil}. 
2)  A post-processing operation is required to find a single maximum of the heatmap. This operation is often
heuristic, and non-differentiable, which precludes end-to-end training-based approaches. 3) The resolution of heatmaps predicted by the FCNs is usually lower than the resolution of the input image. The reduced resolution results in a quantization error and limits the precision of keypoint localization. This quantization error might be ameliorated somewhat by various forms of interpolation, but this makes the framework less differentiable, more complicated and introduces some extra hyper-parameters.

\begin{figure}
	\centering
	\begin{minipage}[t]{0.45\linewidth}
		\centering
		\includegraphics[width=2.1in]{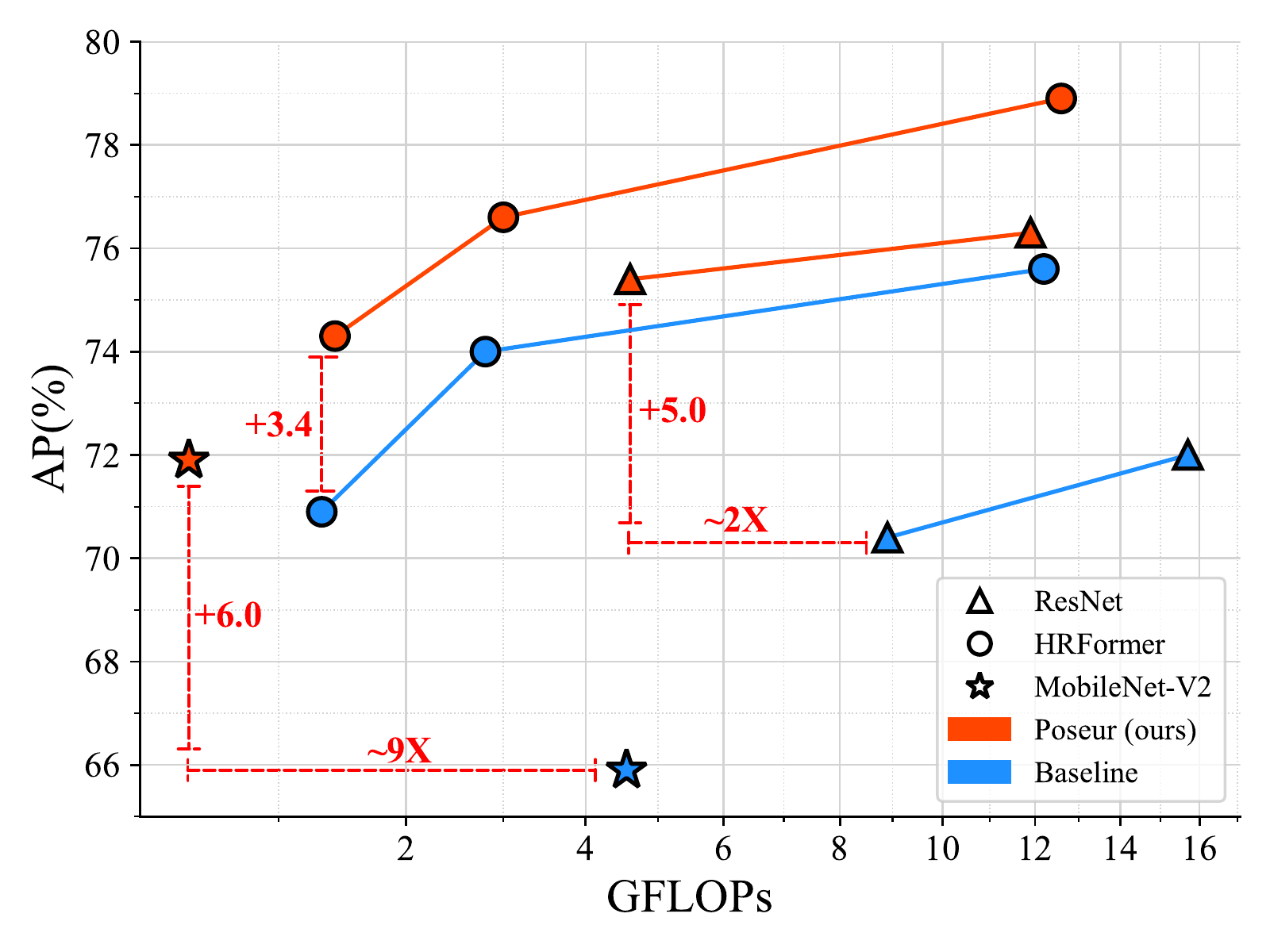}
		\caption{
        \textbf{Comparing the proposed \handle against heatmap-based methods} with various backbone networks on COCO \emph{val.}\ set. Baseline refers to heatmap-based methods. Heatmap-based baseline of MobileNet-V2 and ResNet use the same deconvolutional head as SimpleBaseline~\cite{xiao2018simple}.
        }
        \label{fig:trade_off}
	\end{minipage}
	\hspace{.2em}
	\begin{minipage}[t]{0.45\linewidth}
		\centering
		\includegraphics[width=2.1in]{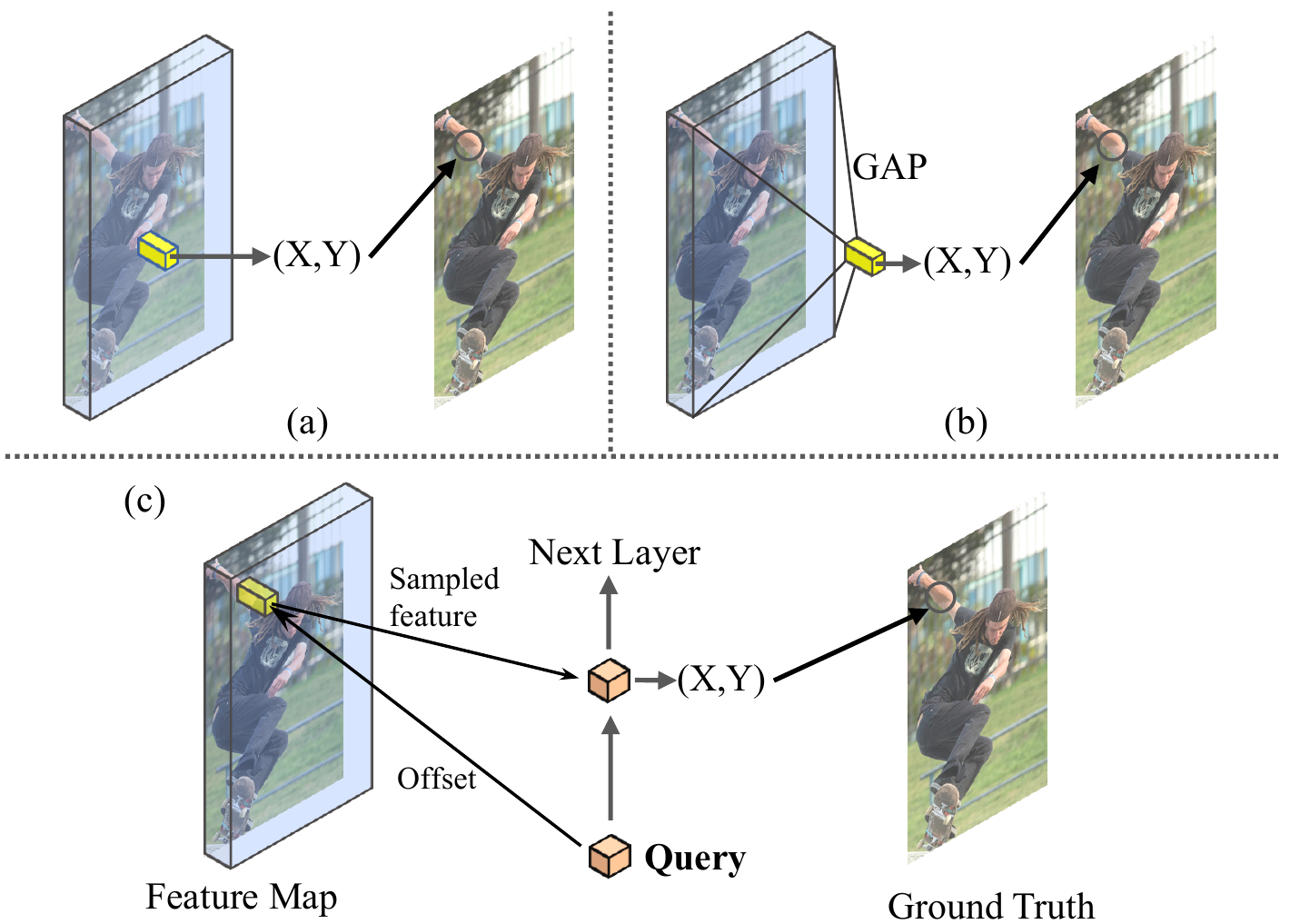}
		\caption{
        \textbf{Comparison of \handle and previous regression-based methods.}  `GAP' indicates global average pooling. (a) shows the feature misalignment issue. (b) shows crucial spatial information is inevitably lost with GAP. We alleviate both issues with the design in (c). \;\;\;\;\;\;
        }
        \label{fig:difference}
	\end{minipage}
\end{figure}

Regression-based methods directly map the input image to the coordinates of body joints, typically using a fully-connected (FC) prediction layer, eliminating the need for heatmaps. The pipeline of regression-based methods is much more {straightforward} than heatmap-based methods, as pose estimation %
is 
naturally
formulated
as a process of predicting a set of 
coordinate values. 
A regression-based approach also alleviates the need for non-maximum suppression, heatmap generation, and quantization compensation, and is inherently end-to-end differentiable.

Regression-based pose-estimation has received less attention than heatmap-based methods due to its inferior performance. 
There are a variety of causes of this performance deficit. First, in order to reduce the number of parameters in the
final FC
prediction layer, models such as DeepPose~\cite{2014deeppose} and RLE~\cite{li2021rle}
employ 
a global average pooling that is applied to reduce the CNN feature map's resolution before the FC layers, as illustrated in \cref{fig:difference}(b).
This global average pooling destroys the spatial structure of the convolutional feature maps, and has a significantly negative impact on performance. Next, as shown in \cref{fig:difference}(a), the convolutional features and predictions of some regression-based models (e.g.  DirectPose~\cite{tian2019directpose} and SPM~\cite{nie2019single}) are misaligned, which consequently reduces localization precision. Lastly, regression-based methods only regress the coordinates of body joints and do not exploit the structured dependency between them~\cite{sun2017compositional}. %

Recently, Transformers have been applied to a range of tasks in computer vision, achieving impressive results~\cite{zhu2020deformable,dosovitskiy2020image,carion2020end}. This, and the fact that transformers were originally designed for sequence-to-sequence tasks, motivated our formulation of single person pose estimation as a sequence prediction problem.  Specifically, we pose the problem as that of predicting a length-$K$ sequence of coordinates, where $K$ is the number of body joints for one person. This leads to a simple and novel regression-based pose estimation framework, that we label as \textbf{\handle}.

As shown in~\cref{fig:framework}, taking as inputs the feature maps of an encoder CNN, the transformer predicts $K$ coordinate pairs. %
In doing so, \handle alleviates the aforementioned difficulties of regression-based methods. First, it does not need global average pooling to reduce feature dimensionality (\cf RLE~\cite{li2021rle}). Second, \handle eliminates the misalignment between the backbone features and predictions with the proposed efficient cross-attention mechanism.
Third, since the self-attention module is applied across the keypoint queries, the transformer naturally captures the structured dependencies among the keypoints.
Lastly, as shown in~\cref{fig:trade_off}, \handle outperforms heatmap-based methods with a variety of backbones. The improvement is more significant for the backbones using low-resolution representation, e.g., MobileNet V2 and ResNet. The results indicate that \handle can be deployed with fast backbones of low-resolution representation without large performance drop, which is difficult to be achieved for heatmap-based methods. We refer readers to \cref{sec:main_result} for more details.

Our main contributions are as follows.
\begin{itemize}

\item  We propose a transformer-based framework (termed \textbf{\handle}) for directly human pose regression, which is lightweight and can work well with the backbones using low-resolution representation. For example, with 49\% fewer FLOPs, ResNet-50 based \handle outperforms the heatmap-based method SimpleBaseline~\cite{xiao2018simple} by $5.0$ AP on the COCO \emph{val} set.
\item \handle significantly improves the performance of regression-based methods, to the point where it is comparable to the state-of-the-art heatmap-based approaches. For example, it improves on the previously best regression-based method (RLE~\cite{li2021rle}) by $4.9$ AP with the ResNet-50 backbone on the COCO \emph{val} set and outperforms the previously best heatmap-based method UDP-Pose~\cite{sun2019deep} by $1.0$ AP with HRNet-W48 on the COCO \emph{test-dev} set.
\item Our proposed framework can be easily extended to an end-to-end pipeline without manual crop operation, for example, we integrate \handle into Mask R-CNN~\cite{he2017mask}, which is end-to-end trainable and can overcome many drawbacks of the heatmap-based methods. In this end-to-end setting, our method outperforms the previously best end-to-end top-down method PointSet Anchor~\cite{wei2020point} by $3.8$ AP with the HRNet-W48 backbone on the COCO \emph{val} set.

\end{itemize}

\section{Related Work}

\noindent\textbf{Heatmap-based pose estimation.}
Heatmap-based 2D pose estimation methods\cite{chen2018cascaded,xiao2018simple,sun2019deep,cai2020learning,li2019rethinking,cao2019openpose,cheng2020higherhrnet,he2017mask,newell2016stacked} estimate per-pixel likelihoods for each keypoint location, and currently dominate in the field of 2D human pose estimation. 
A few works~\cite{newell2016stacked,sun2019deep,cai2020learning} attempt to design powerful backbone networks which can maintain high-resolution feature maps for heatmap supervision.
Another line of works~\cite{sun2017compositional,Zhang_2020_CVPR,huang2020devil} focus on alleviating biased data processing pipeline for heatmap-based methods. Despite the good performance, the heatmap representation bears a few drawbacks in nature, e.g. , non-differentiable decoding pipeline~\cite{sun2018integral,tian2019directpose} and quantization errors~\cite{huang2020devil,Zhang_2020_CVPR} due to the down sampling of feature maps.

\noindent\textbf{Regression-based pose estimation.}
2D human pose estimation is naturally a regression problem\cite{sun2018integral}. However, regression-based methods have historically not been as accurate as heatmap-based methods, and it has received less attention as a result\cite{2014deeppose,sun2018integral,sun2017compositional,carreira2016human,tian2019directpose,nie2019single}. 
 Integral Pose~\cite{sun2018integral} proposes integral regression, which shares the merits of both heatmap representation and regression approaches, to avoid non-differentiable post-processing and quantization error issues. However, integral regression is proven to have an underlying bias compared with direct regression according to~\cite{gu2021removing}.
RLE~\cite{li2021rle} develops a regression-based method using maximum likelihood estimation and flow models. RLE~\cite{li2021rle} is the first to push the performance of the regression-based method to a level comparable with that of the heatmap-based methods. However, it is trained on the backbone that pre-trained by the heatmap loss.

\noindent\textbf{Transformer-based architectures.} %
Transformers have been applied to the pose estimation task with some success. 
TransPose~\cite{yang2021transpose} and HRFormer~\cite{yuan2021hrformer} enhance the backbone via applying the Transformer encoder to the backbone; TokenPose~\cite{li2021tokenpose} designs the pose estimation network in a ViT-style fashion by splitting image into patches and applying class tokens, which makes the pose estimation more explainable. These methods are all heatmap-based and use a heavy transformer encoder to improve the model capacity. In contrast, Poseur is a regression-based method with a lightweight transformer decoder. Thus, Poseur is more computational efficient while can still achieve high performance. 

PRTR~\cite{li2021PRTR} leverage the encoder-decoder structure in transformers to perform pose regression. PRTR is based on DETR~\cite{carion2020end}, i.e., it uses Hungarian matching strategy to find a bipartite matching between non class-specific queries and ground-truth joints.  It brings two issues: 1) heavy computational cost; 2) redundant queries for each instance. In contrast, \handle can alleviate both issues while achieving much higher performance.

\section{Method}

\begin{figure*}[t]
\centering 
\includegraphics[width=0.98\textwidth]{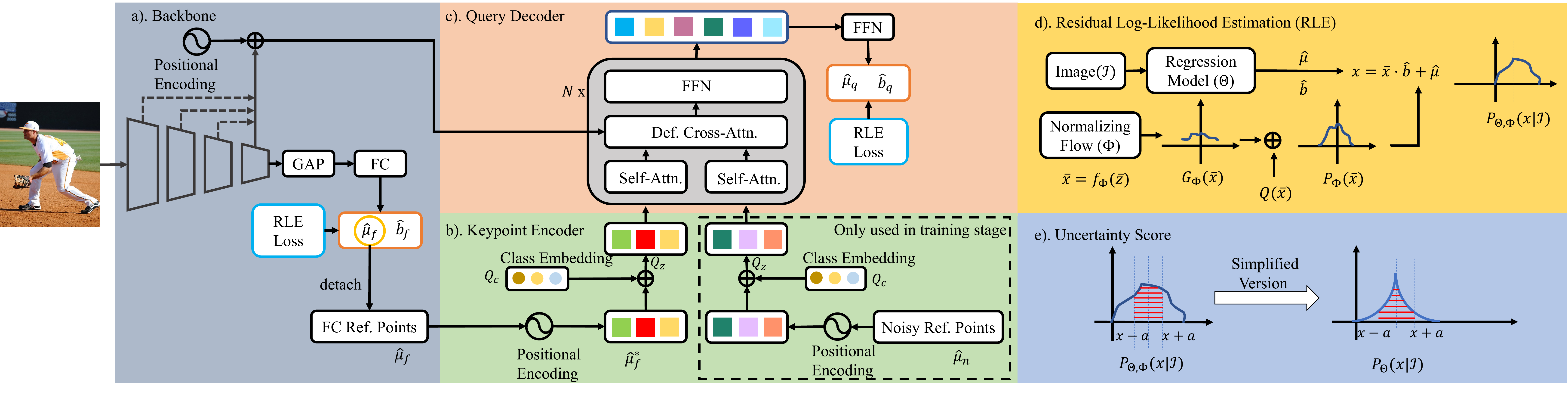}
\caption{\textbf{The architecture of \handle.}
The model directly predicts a sequence of keypoint coordinates in parallel by combining (a) backbone network with (b) keypoint encoder and (c) query decoder. (d) Residual Log-likelihood Estimation~\cite{li2021rle}. (e) The proposed uncertainty score for our method.
}
\label{fig:framework}
\end{figure*}

\subsection{\handle Architecture}\label{sec:Architecture}

Our proposed pose estimator \handle aims to predict $K$ human keypoint coordinates from a cropped single person image.
As shown in \cref{fig:difference}(c), The core idea of our method is to represent human keypoints with queries, \ie, each query corresponds to a human keypoint. The queries are input to the deformable attention module~\cite{zhu2020deformable}, which adaptively attends to the image features that most relevant to the query/keypoint. In this way, the information about a specific keypoint can be summarized and encoded into a single query, which is used to regress the keypoint coordinate later. As such, the issue of losing spatial information caused by the global average pooling in RLE~\cite{li2021human} (As shown in \cref{fig:difference}(b)) is well addressed.

Specifically, in \handle framework (shown in \cref{fig:framework}), two main components are added upon the backbone: a keypoint encoder and a query decoder. An input image is first encoded as dense feature maps with the backbone, which are followed by an FC layer to predict the rough keypoint coordinates, used as a set of rough proposals.
We denote the proposal coordinates as $\KPF \in \R^{K \times 2}$. Then, those proposals are used to initialize the keypoint-specific query $\mathbf{Q} \in \R^{K \times C}$ (where $C$ is the embedding dimension) in the keypoint encoder. Finally, the feature maps from the backbone and $\mathbf{Q}$ are sent into the query decoder to obtain the final features for the keypoints, each of which is sent into a linear layer to predict the corresponding keypoint coordinates.
In addition, unlike previous methods simply regressing the keypoint coordinates and applying the $L_1$ loss %
for supervision, 
\handle, following RLE ~\cite{li2021human}, predicts a probability distribution reflecting the probability of the ground truth appearing in each location and supervise the network by maximum the probability on the ground truth location. Specifically, a location parameter $\KPQ$ and a scale parameter $\KBQ$ are predicted by \handle ($\Theta$) for shifting and scaling the distribution generated by a flow model $\Phi$ (refer to~\cref{sec:loss}). $\KPQ$ is the center of the distribution and can be regarded as the predicted keypoint coordinates.

\noindent\textbf{Backbone.} Our method is applicable to both CNN (e.g. ResNet~\cite{he2016deep}, HRNet~\cite{sun2019deep}) and transformer backbones (e.g. HRFormer~\cite{yuan2021hrformer}).  Given the backbone, multi-level feature maps are extracted and then fed into the query decoder. At the same time, a global average pooling operation is conducted in the last stage of the backbone and followed by an FC layer to regress the coarse keypoint coordinates $\KPF$ (normalized in $[0,1]$) and the corresponding scale parameter $\KBF$, supervised by Residual Log-Likelihood Estimation (RLE) process introduced in~\cref{sec:loss}.

\noindent\textbf{Keypoint encoder.} 
The keypoint encoder is used to initialize each query $\mathbf{Q}$ for the query decoder. For initializing the query better, two keypoints' attributes, location and category, are encoded into the query in the keypoint encoder.
Specifically, first, for location attribute, we encode the rough x-y keypoint coordinates $\KPF$ with the fixed positional encodings, transforming the x-y coordinates to the sine-cosine positional embedding following ~\cite{vaswani2017attention}. The obtained tensor is denoted by $\KPF^* \in \R^{K \times C}$; second, for the category attribute, $K$ learnable vectors $\Qc \in \R^{K \times C}$, called class embedding, is used to represent K different categories separately. Finally, the initial queries $\QZ \in \R^{K \times C}$ are generated by fusing the location and category attribute through element-wise addition of the positional and class embedding, \ie $\QZ = \Qc + \KPF^*$. 
However, $\KPF$ is just a coarse proposal, which sometimes goes wrong during inference. To make our model more robust for the wrong proposal, we introduce a query augmentation process, named \textit{noisy reference points sampling strategy}, used only during training. The core idea of noisy reference points sampling strategy is to simulate the case that the coarse proposals $\KPF$ goes wrong and force the decoder to located correct keypoint with wrong proposal. Specifically, during training, we construct two types of keypoint queries. The first type of keypoint query is initialized with the proposal $\KPF$; the second type of keypoint query is initialized with normalized random coordinates $\kpn$ (noisy proposal). And then, both of two types query are processed equally in all following training stages. Our experiment shows that training the decoder network with noisy proposal $\kpn$ improves its robustness to errors introduced by coarse proposal $\KPF$ during the inference stage. Note, that during inference randomly initialized keypoint queries are not used.
\noindent\textbf{Query decoder.} 
In query decoder, query and feature map are mainly used to module the relationship between keypoints and input image. As shown in \cref{fig:framework}, the decoder follows the typical transformer decoder paradigm, in which,
there are N identical layers in the decoder, each layer consisting of self-attention, cross-attention and feed-forward networks (FFNs). The query $\mathbf{Q}$ goes through these modules sequentially and generates an updated $\mathbf{Q}$ as the input to the next layer. As in DETR\cite{carion2020end}, the self-attention and FFNs are a multi-head self-attention~\cite{vaswani2017attention} module and MLPs, respectively. 
For the cross-attention networks, we propose an efficient multi-scale deformable attention (EMSDA) module, based on MSDA proposed by Deformable DETR~\cite{zhu2020deformable}. Similar to MSDA, in EMSDA, each query learns to sample relevant features from the feature maps by given the sampling offset around a reference point (a pair of coordinates, and which will be introduced later); and then, the sampled features are summarized by the attention mechanism to update the query. Different from MSDA, which applies a linear layer to the entire feature maps and thus is less efficient, we found that it is enough to only apply the linear layer to the sampled features after bilinear interpolation. Experiments show that the latter can have a similar performance while being much more efficient. Specifically, EMSDA can be written as 
\begin{equation}
\small
\begin{aligned}
    &\operatorname{EMSDA}(\Q_q,\hat{\p}_q,\{\x^l\}_{l=1}^L) = \operatorname{Concat}(\text{head}_{\mathrm{1}},\ldots,\text{head}_{\mathrm{M}})\W^o \\
    &\text{where}\ \ \text{head}_{\mathrm{i}} = (\sum\limits_{l=1}^{L}{\sum\limits_{s=1}^{S}{\A_{i,l,q,s} \cdot \x^l(\phi_l(\hat{\p}_q)+\Delta{p_{i,l,q,s}})}})\W^v_i , \\
\end{aligned}
\end{equation}
where $\Q_q \in \R^{C}$, $\hat{\p}_q \in \R^{2}$ and $\{\x^l\}_{l=1}^L$ are the $q$-th input query vector, the reference point offset of $q$-th query and $l$-th level of feature maps from the backbone; the dimension of each feature vector in $\x$ is $C$. $\text{head}_i$ represents $i$-th attention head. $L$, $M$ and $S$ represent the number of feature map levels used in the decoder, the number of attention heads and the number of sampling points on each level feature map, respectively. $\A_{i,l,q,s} \in \R^{1}$ and $\Delta{p_{i,l,q,s}} \in \R^{2}$ represent the attention weights and the sampling offsets of the $i$-th head, $l$-th level, $q$-th query and $s$-th sampling point, respectively; The query feature $\Q_q$ is fed to a linear projection to generate $\A_{i,l,q,s}$ and $\Delta{p_{i,l,q,s}}$.
$\A_{i,l,q,s}$ satisfies the limitation, $\sum\limits_{l=1}^{L}{\sum\limits_{s=1}^{S}{\A_{i,l,q,s}}}=1$. $\phi_l(\cdot)$ is the function transforming the $\hat{\p}_q$ to the coordinate system of the $l$-th level features. $\x^l(\phi_l(\hat{\p}_q)+\Delta{p_{i,l,q,s}})$ represents sampling the feature vector located in offset $(\phi_l(\hat{\p}_q)+\Delta{p_{i,l,q,s}})$ on the feature map $\x^l$ by bilinear interpolation. $\W^o \in \R^{C \times C}$ and $\W^v_i \in \R^{C \times (C/M)}$ are two groups of trainable weights. The reference point $\hat{\p}_q$ will be updated at the end of each decoder layer by applying a linear layer on $\Q_q$. Note, the FC output $\KPF$ is leveraged as reference point for the initial query $\QZ$. For more details and computational complexity, we refer readers to our supplementary material.

To sum up, the relations between different keypoints are modeled through a self-attention module, and the relations between the input image and keypoints are modeled through EMSDA module. Notably, the problem of feature misalignment in fully-connected regression is solved by EMSDA. %

\subsection{Training Targets and Loss Functions}\label{sec:loss}
Following RLE~\cite{li2021rle}, we calculate a probability distribution $P_{{\BTheta}, {\BPhi}}(\x | \mathcal{I})$ reflecting the probability of the ground truth appearing in the location $\x$ conditioning on the input image $\mathcal{I}$, where $\BTheta$ is the parameters of \handle and $\BPhi$ is the parameters of a flow model. As shown in~\cref{fig:framework}(d), The flow model $f_{\phi}$ is leveraged to reflect the deviation of the output from the ground truth $\loc_g$ by mapping a initial distribution $\bar{\textbf{z}} \sim \mathcal{N}(0, \mathbf{I})$ to a zero-mean complex distribution $\bar{\textbf{x}} \sim G_{\phi}(\bar{\textbf{x}})$. 
Then $P_{\phi}(\bar{\x})$ is obtained by adding a zero-mean Laplace distribution $L(\bar{\x})$ to $G(\bar{\x})$.
The regression model $\BTheta$ predictions the center $\hat{\loc}$, and scale $\hat{\b}$ of the distribution. Finally, the distribution $P_{{\BTheta}, {\BPhi}}(\x | \mathcal{I})$ is built upon $P_{\phi}(\bar{\x})$ by shifting and rescaling $\bar{\x}$ into $\x$, where $\x = \bar{\x} \cdot \hat{\boldsymbol{\sigma}} + \hat{\loc}$. We refer readers to~\cite{li2021rle} for
more details.

Different from RLE~\cite{li2021rle}, we only use the proposal $(\hat{\loc}_{f}, \hat{\boldsymbol{b}}_{f})$ for coarse prediction. This prediction is then updated by the query-based approach described above to generate an improved estimate $(\hat{\loc}_{q}, \hat{\boldsymbol{b}}_{q})$. Both coarse proposal $(\hat{\loc}_{f}, \hat{\boldsymbol{b}}_{f})$ and query decoder preditions $(\hat{\loc}_{q}, \hat{\boldsymbol{b}}_{q})$ are supervised with the maximum likelihood estimation (MLE) process. The learning process of MLE optimizes the model parameters so as to make the observed ground truth $\loc_{g}$ most probable. The loss function of FC predictions $(\hat{\loc}_{f}, \hat{\boldsymbol{b}}_{f})$ can be defined as:
\begin{equation}\label{equation:fc_loss}
\begin{aligned}
    \mathcal{L}_{rle}^{fc} = - \log P_{{\BTheta}_{f},{\BPhi}_{f}}(\x | \mathcal{I}) \Big|_{\x = \loc_g},
\end{aligned}
\end{equation}
where $\BTheta_{f}$ and $\BPhi_{f}$ are the parameters of the backbone and flow model, respectively.
Similarly, the loss of distribution associated with query decoder preditions $(\hat{\loc}_{q}, \hat{\boldsymbol{b}}_{q})$ can be defined as:
\begin{equation}\label{equation:loss func}
    \mathcal{L}_{rle}^{dec} = - \log P_{{\BTheta}_{q},{\BPhi}_{q}}(\x | \mathcal{I}) \Big|_{\x = \loc_g},
\end{equation}
where $\BTheta_{q}$ and $\BPhi_{q}$ are the parameters of the query decoder and another flow model, respectively.
Finally, we sum the two loss functions to obtain the total loss:
\begin{equation}\label{equation:loss func}
    \mathcal{L}_{total} = \mathcal{L}_{rle}^{fc} + \lambda \mathcal{L}_{rle}^{dec},
\end{equation}
where $\lambda$ is a constant and used to balance the two losses. We set $\lambda = 1$ by default.

\subsection{Inference}\label{sec:score}
\noindent\textbf{Inference pipeline.} 
During the inference stage, \handle predicts the $(\hat{\loc}_{q}, \hat{\b}_{q})$ for each keypoint as mentioned; $\hat{\loc}_{q}$ is taken as the predicted keypoint coordinates and $\hat{\boldsymbol{b}}_{q}$ is used to calculate the keypoint confidence score.

\noindent\textbf{Prediction uncertainty estimation.} 
For heatmap-based methods,  e.g.  SimpleBaseline~\cite{xiao2018simple}, the prediction score of each keypoint is combined with a bounding box score to enhance the final human instance score:
\begin{equation}
\begin{aligned}
    \boldsymbol{s}^{inst} = \boldsymbol{s}^{bbox}\frac{\sum_{i=1}^{k}\boldsymbol{s}^{kp}_{i}}{K},
\end{aligned}
\end{equation}
where $\boldsymbol{s}^{inst}$ is the final prediction score of the instance; $\boldsymbol{s}^{bbox}$ is the bounding box score predicted by the person detector, $\boldsymbol{s}_{i}^{kp}$ is the $i$-th keypoint score predicted by the keypoint detector and $K$ is the total keypoint number of each human. 
Most previous regression-based methods~\cite{2014deeppose,sun2018integral} ignore the importance of the keypoint score.
As a result, compared to heatmap based methods, regression methods typically achieve higher recall but lower precision.
Given the same well-trained \handle model, adding the keypoint score brings $4.7$ AP improvement ($74.7$ AP vs. $70.0$ AP) due to the significantly reduced number of false positives, and both of the models achieve almost the same average recall (AR).

Our model predicts a probability distribution over the image coordinates for each human keypoint. 
We define the $i$-th keypoint prediction score $\boldsymbol{s}^{kp}_{i}$ to be the probability of the keypoint falling into the region ($[\hat{\loc}_{i}-\a,\hat{\loc}_{i}+\a]$) near the prediction coordinate $\hat{\loc}_{i}$, \ie
\begin{equation}
\begin{aligned}
    \boldsymbol{s}_{i}^{kp} = \int_{\hat{\loc}_{i}-\a}^{\hat{\loc}_{i}+\a} P_{{\BTheta}_{q},{\BPhi}_{q}}(\x | \mathcal{I}) dx,
\end{aligned}
\end{equation}
where $\a$ is a hyperparameter that controls the size of the $\loc$-adjacent interval, and $\hat{\loc}_{i}$ are the coordinates of the corresponding keypoint predicted by \handle.
In practice, running the normalization flow model during the inference stage would add more computational cost. We found that comparable performance can be achieved by shifting and re-scaling the zero-mean Laplace distribution $L(\bar{\x})$ with query decoder predictions $(\hat{\loc}_{q}, \hat{\boldsymbol{b}}_{q})$. So the probability density function can be rewritten as:
\begin{equation}\label{equation:score}
\begin{aligned}
    P_{{\BTheta}_{q},{\BPhi}_{q}}(\x | \mathcal{I}) \approx f(\x|\hat{\loc}_{i}, \hat{\boldsymbol{b}}_{i}) = \frac{1}{2\hat{\boldsymbol{b}}_{i}}\exp{(-\frac{|\x-\hat{\loc}_{i}|}{\hat{\boldsymbol{b}}_{i}})},
\end{aligned}
\end{equation}
where $\hat{\loc}_{i}$ is the center of the Laplacian distribution and the predicted keypoint coordinates, and $\hat{\boldsymbol{b}}_{i}$ is the scale parameter predicted by \handle. Finally, $\boldsymbol{s}_{i}^{kp}$ can be written as:
\begin{equation}
\begin{aligned}
    \boldsymbol{s}_{i}^{kp} = \int_{\hat{\loc}_{i}-\a}^{\hat{\loc}_{i}+\a} f(\x|\hat{\loc}_{i}, \hat{\boldsymbol{b}}_{i}) dx = 1 - \exp{(-\frac{\a}{\hat{\boldsymbol{b}}_{i}})}. \label{eq8}
\end{aligned}
\end{equation}
Note that the score $\boldsymbol{s}_{i}^{kp}$ on x-axis and y-axis will be calculated separately and then merged by a multiplication operation.

\section{Experiments}

\subsection{Implementation Details}
\noindent\textbf{Datasets.} 
Our experiments are mainly conducted on COCO2017 Keypoint Detection\cite{wang2018mscoco} benchmark, which contains about $250K$ person instances with 17 keypoints. 
We report results on the \emph{val} set for ablation studies and compare with other state-of-the-art methods on both of the \emph{val} set and \emph{test-dev} sets. The Average Precision (AP) based on Object Keypoint Similarity (OKS) is employed as the evaluation metric on COCO dataset. We also conduct experiments on MPII~\cite{mpii} dataset with Percentage of Correct Keypoint (PCK) as evaluation metric.

\noindent\textbf{Model settings.} 
Unless specified, ResNet-50\cite{he2016deep} is used as the backbone in ablation study. 
The size of input image is $256\times192$.
The weights pre-trained on ImageNet\cite{deng2009imagenet} are used to initialize the ResNet backbone. The rest parts of our network are initialized with random parameters. All the decoder embedding size is set as 256; 3 decoder layers are used by default.

\noindent\textbf{Training.}  All the models are trained with batch size 256 (batch size 128 for HRFormer-B due to the limited GPU memory), and are optimized by AdamW\cite{loshchilov2017decoupled} with a base learning rate of $1\times10^{-3}$ decreased to $1\times10^{-4}$ and $1\times10^{-5}$ at the 255-th epoch and 310-th epoch and ended at the 325-th epoch; 
$\beta_1$ and $\beta_2$ are set to $0.9$ and $0.999$, respectively;
Weight decay is set to $10^{-4}$. 
Following Deformable DETR~\cite{zhu2020deformable}, the learning rate of the the linear projections for sampling offsets and reference points are multiplied by a factor of $0.1$. Following RLE~\cite{li2021rle}, we adopt RealNVP~\cite{realnvp} as the flow model. Other settings follow that of mmpose~\cite{mmpose2020}. For HRNet-W48 and HRFormer-B, cutout~\cite{devries2017improved} and color jitter augmentation are applied to avoid over-fitting.

\noindent\textbf{Inference. }
Following conventional settings, we use the same person detector as in SimpleBaseline~\cite{xiao2018simple} for COCO evaluation.
According to the bounding box generated by the person detector, the single person image patch is cropped out from the original image and resized to a fix resolution, e.g.  $256\times192$. The flow model is removed during the inference. We set $\a=0.2$ in~\cref{equation:score} by default.
\subsection{Ablation Study}

\noindent\textbf{Initialization of keypoint queries.} 
We conduct experiments to verify the impact of initialization of keypoint queries. Deformable DETR~\cite{zhu2020deformable} introduces reference points that represent the location information of object queries. In their paper, reference points are 2-d tensors predicted from the 256-d object queries via a linear projection. We set this configuration as our baseline model. As shown in ~\cref{tab:source_of_init_ref_points}, the baseline model achieves 72.3 AP with 3 decoder layers, which is $0.6$ AP lower than keypoint queries which initialized from coarse proposal $\KPF$.
This indicates that coarse proposal $\KPF$ provide a good initialization for the keypoint queries.

\setlength{\tabcolsep}{0.3pt} %
\begin{table*}[t]
    \caption{Ablation of proposed \handle on COCO \texttt{val2017} split. ``Ours": Using the fully convolutional layer at the end of backbone to regress the coarse proposal $\KPF$; ``Noisy Reference Points": applying the noisy reference points sampling strategy in the keypoint encoder; ``Res-$i$": $i$-th level feature map of ResNet; ``$N_d$": the number of decoder layers}
	\begin{center}
		\subfloat[Varying Initial Reference Points Methods\label{tab:source_of_init_ref_points}]{
            \begin{tabular}{c|c}
                \hline
                Initial Ref. Points   & {AP} \\
                \Xhline{2\arrayrulewidth}
                Def. DETR\cite{zhu2020deformable}  & 72.3 \\
                Ours   & 72.9 \\ %
                \hline
            \end{tabular}
		}
		\hspace{1em}
		\subfloat[Varying the Noisy Reference Points \label{tab:cls_embed_noisy_sampling_robustness}]
		{
                \begin{tabular}{c|c}
                    \hline
                    Noisy Ref. Points & {AP} \\
                    \Xhline{2\arrayrulewidth}
                    \xmark  & 73.7 \\
                    \cmark & 74.3 \\
                    \hline
                \end{tabular}
		}
		\hspace{1em}
		\subfloat[Varying the Uncertainty Estimation \label{tab:score_estimation}]
		{
                    \begin{tabular}{c|c}
                        \hline
                        Uncertainty Esti.   & {AP} \\
                        \Xhline{2\arrayrulewidth}
                        RLE~\cite{li2021rle}   & 73.6 \\
                        Ours  & 74.7 \\
                        \hline
                    \end{tabular}
		}
		\hspace{1em}
		\subfloat[Varying the scale levels of input feature map for decoder \label{tab:feature_levels}]
		{
            		\begin{tabular}{cccc|cc|c}
                    \hline
                    Res2\;        & Res3\;       & Res4\;       & Res5\;        & Params & GFLOPs   & {AP} \\
                    \Xhline{2\arrayrulewidth}
                                &            &            & \cmark & 28.3M & 4.12 & 73.7 \\
                                &            & \cmark & \cmark & 28.7M & 4.18 & 74.2 \\
                                & \cmark & \cmark & \cmark & 28.9M & 4.28 & 74.4 \\
                    \cmark & \cmark & \cmark & \cmark & 29.0M & 4.48 & 74.7 \\
                    \hline
                    \end{tabular}
		}
		\hspace{0.1em}
		\subfloat[Varying the numbers of decoder layers \label{table: num_dec_layers}]
		{
                	\begin{tabular}{c|cc|l|ll}
                	\hline
                	$N_d$ & Params & GFLOPs & AP & AP$_{50}$ & AP$_{75}$ \\
                	\Xhline{2\arrayrulewidth}
                	3 & 28.8M & 4.48 & 74.7  & 90.2 & 81.6 \\
                	4 & 30.2M & 4.51 & 75.3  & 90.5 & 82.1 \\
                	5 & 31.6M & 4.54 & 75.4  & 90.3 & 82.2 \\
                	6 & 33.1M & 4.57 & 75.4  & 90.5 & 82.2 \\
                	\hline
                	\end{tabular}
		}
	\end{center}
\end{table*}
\setlength{\tabcolsep}{0.4pt}

\noindent\textbf{Noisy reference points sampling strategy.} As mentioned in~\cref{sec:Architecture}, we apply the noisy reference points sampling strategy during the training. To validate its effectiveness, we perform ablation experiment on COCO, as show in ~\cref{tab:cls_embed_noisy_sampling_robustness}. The experiment result shows that the noisy reference points sampling strategy can improve the accuracy by 0.6 AP without adding any extra computational cost during inference.

\noindent\textbf{Varying the levels of feature map.} 
We explore the impact of feeding different levels of backbone features into the proposed query decoder. As shown in~\cref{tab:feature_levels}, the performance grows consistently with more levels of feature maps, \textit{e.g.},   73.7 AP, %
74.2 AP,  %
74.4 AP,
74.7 AP %
for 
1, 2, 3, 4 levels of feature maps, respectively.

\noindent\textbf{Uncertainty estimation.} 
As mentioned in ~\cref{sec:score}, we redesign the prediction confidence score proposed in~\cite{li2021rle}. To study the effectiveness of the proposed score $\boldsymbol{s}^{kp}$, we compare it with predictions without re-score and predictions with RLE score~\cite{li2021rle} using the same model. As shown in~\cref{tab:score_estimation}, the proposed method brings significant improvement (4.7 AP) to the model without uncertainty estimation, and outperforms the RLE score~\cite{li2021rle} by 1.0 AP.

\noindent\textbf{Varying decoder layers.} Here we study the effect of query decoder's depth.
Specifically, we conduct experiments by varying the number of decoder layers in Transformer decoder. 
As shown in ~\cref{table: num_dec_layers}, the performance grows at the first three layers and saturates at the sixth decoder layer. 

\begin{table*}[t] %
    \caption{Comparison with heatmap methods by varying the backbone and the input resolution on the COCO \emph{val} set. ``SimBa": SimpleBaseline~\cite{xiao2018simple}. For (a), the input resolution is 256$\times$192 and the number of decoder layers is 5. For (b), we use ResNet-50 as backbone and the number of decoder layers is 3.}
	\begin{center}
		\subfloat[Varying the backbone\label{tab:com_hr_sb_tf}]{
            \begin{tabular}{c|c|c|l}
                \hline
                {Method} & {Backbone}  & GFLOPs & {AP}   \\
                \Xhline{2\arrayrulewidth}
                SimBa.  & MobileNet-V2   &4.55  & 65.9   \\
                \handle &MobileNet-V2    &0.52  & 71.9   \\
                \hline
                SimBa.  & ResNet-50      &8.27  & 72.4   \\
                \handle & ResNet-50      &4.54  & 75.4   \\
                \hline
                HRNet   & HRNet-W32      &7.68  &75.0 \\
                \handle & HRNet-W32      &7.95  & 76.9  \\ 
                \hline
        
            \end{tabular}
		}
		\subfloat[Varying the input resolution \label{tab:input_size}]
		{
            \begin{tabular}{c|cccl}
                \hline
                Method & Input size & Params & GFLOPs  & AP \\
                \Xhline{2\arrayrulewidth}
                SimBa.~\cite{xiao2018simple} & 64$\times$64 & 34.0M & 0.69 & 31.4  \\
                \handle  & 64$\times$64 &28.8M & 0.49 & \textbf{47.9}   \\
                \hline
                SimBa.~\cite{xiao2018simple}  & 128$\times$128 & 34.0M &2.76 &59.3  \\
                \handle & 128$\times$128 & 28.8M & 1.55 & \textbf{67.1}  \\
                \hline
                SimBa.~\cite{xiao2018simple}  & 256$\times$192 & 34.0M & 8.26 & 71.0 \\
                \handle  & 256$\times$192 & 28.8M & 4.48 & \textbf{74.7}  \\
                \hline
            \end{tabular}
		}
	\end{center}
\end{table*}

\noindent\textbf{Varying the input size.} We conduct experiments to explore the robust of \handle under different input resolutions.~\cref{tab:input_size} compares \handle with SimpleBaseline, showing that our method consistently outperforms SimpleBaseline in all input sizes.
The results also indicate that heatmap-based method suffers larger performance drop with the low-resolution input. For example, the proposed method outperforms SimpleBaseline by 14.6 AP in 64$\times$64 input resolution.

\subsection{Extensions: End-to-End Pose Estimation}
Our framework can easily extend to end-to-end human pose estimation, \ie, detecting multi-person poses without the manual crop operation. With \handle as a plug-and-play scheme, end-to-end top-down keypoint detectors can obtain additional improvement. Here, we take Mask-RCNN as example to show the superiority of our method. The original keypoint head of Mask R-CNN is stacked 8 convolutional layers, followed by a deconv layer and 2× bilinear upscaling, producing an output resolution of 56$\times$56. We replace the deconv layer by an average pooling layer and an FC layer like~\cite{li2021rle}. The output of the FC layer is used to produce initial coarse proposal $\KPF$. Then coarse proposal $\KPF$ is feed into the keypoint encoder and query decoder as described in \cref{sec:Architecture}. We randomly sample 600 queries per image for training efficiency. Note that we conduct EMSDA on multi-scale backbone feature maps, rather than on ROI features. The output of FC layer and transformer decoder are both supervised with RLE loss~\cite{li2021rle}.
We perform scale jittering~\cite{ghiasi2021simple} with random crops during training.
We train the entire network for 180,000 iterations, with a batchsize of 32 in total. Other parameters are the same as the Detectron2~\cite{wu2019detectron2}.
As shown in~\cref{tab:maskrcnn}, \handle outperforms the heatmap-based Mask R-CNN with ResNet 101 by 1.9 AP. \handle outperforms the state-of-the-art regression-based method, PointSet Anchor with HRNet-W48 by $3.8$ AP.

\begin{table}[t]
	\centering
	\begin{minipage}[t]{0.52\linewidth}
		\centering
	    \caption{Comparison with \textbf{end-to-end top-down methods} on the COCO \emph{val} set. 
	    $^\dag$ denote flipping and multi-sacle testing. Reg: regression-based approach; HM: heatmap-based approach}\label{tab:maskrcnn}
	    \resizebox{1\linewidth}{!}{
		\begin{tabular}{c|c|c|ccc}
        \hline
        Method & Backbone & Type & AP & AP$_{50}$ & AP$_{75}$ \\
        \Xhline{2\arrayrulewidth}
        PRTR~\cite{li2021PRTR} & HRNet-W48 & Reg. & 64.9 & 87.0 & 71.7 \\
        Mask R-CNN~\cite{he2017mask} & ResNet-101 & HM. & 66.0 & {86.9} & 71.5 \\
        Mask R-CNN + RLE~\cite{li2021rle} & ResNet-101 & Reg. & {66.7} & 86.7 & {72.6} \\
        PointSet Anchor$^\dag$~\cite{wei2020point} & HRNet-W48 & Reg. & 67.0 & 87.3 & 73.5 \\
        {Mask R-CNN + \handle} & ResNet-101 & Reg. & {68.6} & {87.5} & {74.8} \\
        {Mask R-CNN + \handle} & HRNet-W48 & Reg. & {70.1} & \textbf{88.0} & {76.5} \\
        {Mask R-CNN + \handle$^\dag$} & HRNet-W48 & Reg. & \textbf{70.8} & {87.9} & \textbf{77.0} \\
        \hline
        \end{tabular}}
	\end{minipage}
	\hspace{.2em}
	\begin{minipage}[t]{0.4\linewidth}
		\centering
		\caption{Comparisons on MPII validation set (PCKh@0.5). SimBa: SimpleBaseline~\cite{xiao2018simple}. Reg: regression-base approach; HM: heatmap-based approach}\label{tab:mpii}
		\resizebox{1\linewidth}{!}{
    	\begin{tabular}{l|c|c|c}
    	\hline
    	Method & Backbone& Type & Mean \\
        \Xhline{2\arrayrulewidth}
        SimBa.~\cite{xiao2018simple}        & ResNet-152    &HM.& $89.6$  \\
        HRNet~\cite{sun2019deep}            & HRNet-W32     &HM.& $90.1$   \\
        TokenPose~\cite{li2021tokenpose}    & L/D$24$       &HM.& ${90.2}$ \\
        Integral~\cite{sun2018integral}     & ResNet-101    &Reg.&87.3 \\
        PRTR~\cite{li2021PRTR}              & HRNet-W32     &Reg.&89.5 \\
        \handle                             & HRNet-W32     &Reg.& \textbf{90.5} \\
        \hline
	    \end{tabular}}
	\end{minipage}
\end{table}

\begin{table}[!t]
    \small
    \centering
	\caption{\textbf{Comparisons with state-of-the-art methods} on the COCO \emph{val} set.  Input size and the GFLOPs are calculated under top-down single person pose estimation setting. Unless specified, the number of decoder layers is set to 6. ``3 Dec.": three decoder layers.
	}
    \resizebox{0.9\columnwidth}{!}{
		\smallskip\begin{tabular}{r |c|c|c|c|c|c|c|c}
		\hline
	    Method                                   & Backbone / Type  & Input Size   & GFLOPs   & AP$^{kp}$ & AP$^{kp}_{50}$ & AP$^{kp}_{75}$ & AP$^{kp}_{M}$ & AP$^{kp}_{L}$ \\
	    \hline
	    \multicolumn{5}{c}{\textbf{Heatmap-based methods}} \\
	    \hline
        SimBa.~\cite{xiao2018simple}  & ResNet-50        & $256\times192$ &  8.9         & 70.4 & 88.6 & 78.3 & 67.1 & 77.2 \\
        SimBa.~\cite{xiao2018simple}  & ResNet-152       & $256\times192$ &  15.7        & 72.0 & 89.3 & 79.8 & 68.7 & 78.9 \\
        HRNet~\cite{sun2019deep}          & HRNet-W32        & $256\times192$ &  7.1         & 74.4 & 90.5 & 81.9 & 70.8 & 81.0 \\
        HRNet~\cite{sun2019deep}          & HRNet-W48        & $384\times288$ &  32.9        & 76.3 & 90.8 & 82.9 & 72.3 & 83.4 \\
        TransPose~\cite{yang2021transpose}       & H-A6             & $256\times192$ &  21.8        & 75.8 &   -  &   -  &   -  &   -  \\
        TokenPose~\cite{li2021tokenpose}         & S-V2             & $256\times192$ &  11.6        & 73.5 & 89.4 & 80.3 & 69.8 & 80.5 \\
        TokenPose~\cite{li2021tokenpose}         & B                & $256\times192$ &  5.7         & 74.7 & 89.8 & 81.4 & 71.3 & 81.4 \\
        TokenPose~\cite{li2021tokenpose}         & L/D6             & $256\times192$ &  9.1         & 75.4 & 90.0 & 81.8 & 71.8 & 82.4 \\
        TokenPose~\cite{li2021tokenpose}         & L/D24            & $256\times192$ &  11.0        & 75.8 & 90.3 & 82.5 & 72.3 & 82.7 \\
        HRFormer~\cite{yuan2021hrformer}       & HRFormer-T       & $256\times192$ &  1.3         & 70.9 & 89.0 & 78.4 & 67.2 & 77.8 \\
        HRFormer~\cite{yuan2021hrformer}       & HRFormer-S       & $256\times192$ &  2.8         & 74.0 & 90.2 & 81.2 & 70.4 & 80.7 \\
        HRFormer~\cite{yuan2021hrformer}       & HRFormer-B       & $256\times192$ &  12.2        & 75.6 & 90.8 & 82.8 & 71.7 & 82.6 \\
        HRFormer~\cite{yuan2021hrformer}       & HRFormer-B       & $384\times288$ &  26.8        & 77.2 & 91.0 & 83.6 & 73.2 & 84.2 \\
        UDP-Pose~\cite{huang2020devil}           & HRNet-W32       & $256\times192$ &  7.2        & 76.8 & 91.9 & 83.7 & 73.1 & 83.3 \\
        UDP-Pose~\cite{huang2020devil}           & HRNet-W48       & $384\times288$ &  33.0        & 77.8 & 92.0 & 84.3 & 74.2 & 84.5 \\
        \hline
        \multicolumn{5}{c}{\textbf{Regression-based methods}} \\
	    \hline
	    PRTR~\cite{li2021PRTR}                      & ResNet-50  & $384\times288$ & 11.0     & 68.2 & 88.2 & 75.2 & 63.2 & 76.2 \\
	    PRTR~\cite{li2021PRTR}                      & HRNet-W32  & $384\times288$ & 21.6     & 73.1 & 89.4 & 79.8 & 68.8 & 80.4 \\
        PRTR~\cite{li2021PRTR}                      & HRNet-W32  & $512\times384$ & 37.8     & 73.3 & 89.2 & 79.9 & 69.0 & 80.9 \\
		RLE~\cite{li2021human}                      & ResNet-50         & $256\times192$ & 4.0      & 70.5 & 88.5 & 77.4 & -    &  -   \\
		RLE~\cite{li2021human}                      & HRNet-W32 & $256\times192$ & ~7.1 & 74.3 & 89.7& 80.8 & -   & - \\
		Ours                                        & MobileNet-v2         & $256\times192$ &  0.5   & 71.9 & 88.9 & 78.6 & 65.2 & 74.3 \\
		Ours                                          & ResNet-50         & $256\times192$ &  4.6     & 75.4 & 90.5 & 82.2 & 68.1 & 78.6 \\ %
		Ours                                          & ResNet-152      & $256\times192$ &  11.9    & 76.3 & 91.1 & 83.3 & 69.1 & 79.5 \\ %
		Ours                                        & HRNet-W32         & $256\times192$ &   7.4   & 76.9 & 91.0 & 83.5 & 70.1 & 79.7 \\ %
		Ours                                        & HRNet-W48         & $384\times288$ &   33.6   & 78.8 & 91.6 & 85.1 & 72.1 & 81.8 \\ %
		Ours (3 Dec.)                               & HRFormer-T         & $256\times192$ &  1.4    & 74.3 & 90.1 & 81.4 & 67.5 & 76.9 \\ %
		Ours (3 Dec.)                               & HRFormer-S         & $256\times192$ &  3.0     & 76.6 & 91.0 & 83.4 & 69.8 & 79.4 \\ %
		Ours                                        & HRFormer-B         & $256\times192$ & 12.6     & 78.9 & 92.0 & 85.7 & 72.3 & 81.7 \\ %
		Ours                                        & HRFormer-B         & $384\times288$ & 27.4     & 79.6 & 92.1 & 85.9 & 72.9 & 82.9 \\ %
    \hline
	\end{tabular}}
	\label{tab:comparisons_with_sota_on_val}
\end{table}

\begin{table}[!t]
    \begin{center}
    \caption{\textbf{Comparison with top-down methods} on the COCO \emph{test-dev} set. The proposed paradigm outpeforms heatmap-based methods in various settings.
    The input resolution of all methods is $384\times288$.}
    \resizebox{0.65\linewidth}{!}
    {%
		\begin{tabular}{l|c|c|c|c|c|c}
		\hline
	    Method & Backbone  & AP$^{kp}$ & AP$^{kp}_{50}$ & AP$^{kp}_{75}$ & AP$^{kp}_{M}$ & AP$^{kp}_{L}$ \\
	    \hline
	    \multicolumn{7}{c}{\textbf{Heatmap-based methods}} \\
	    \hline
        SimBa$^\dag$~\cite{xiao2018simple}          & ResNet-152        & 73.7 & 91.9 & 81.1 & 70.3 & 80.0 \\ %
        HRNet$^\dag$~\cite{sun2019deep}                 & HRNet-W32         & 74.9 & 92.5 & 82.8 & 71.3 & 80.9 \\ %
        HRNet$^\dag$~\cite{sun2019deep}                 & HRNet-W48         & 75.5 & 92.5 & 83.3 & 71.9 & 81.5 \\ %
        TokenPose~\cite{li2021tokenpose}                & L/D24             & 75.9 & 92.3 & 83.4 & 72.2 & 82.1 \\ %
        HRFormer~\cite{yuan2021hrformer}              & HRFormer-B        & 76.2 & 92.7 & 83.8 & 72.5 & 82.3 \\ %
        UDP-Pose~\cite{huang2020devil}                  & HRNet-W48         & 76.5 & 92.7 & 84.0 & 73.0 & 82.4 \\ %
        \hline
        \multicolumn{7}{c}{\textbf{Regression-based methods}} \\
	    \hline
	    PRTR~\cite{li2021PRTR}                          & ResNet-101        & 68.8 & 89.9 & 76.9 & 64.7 & 75.8 \\ %
        PRTR~\cite{li2021PRTR}                          & HRNet-W32         & 71.7 & 90.6 & 79.6 & 67.6 & 78.4 \\ %
        RLE~\cite{li2021human}                          & ResNet-152        & 74.2 & 91.5 & 81.9 & 71.2 & 79.3 \\ %
        RLE~\cite{li2021human}                          & HRNet-W48 & 75.7 & 92.3 & 82.9 & 72.3 & 81.3 \\ %
		Ours (6 Dec.)                                  & HRNet-W48         & 77.6 & 92.9 & 85.0 & 74.4 & 82.7 \\
		Ours (6 Dec.)                                 & HRFormer-B        & \textbf{78.3} & \textbf{93.5} & \textbf{85.9} & \textbf{75.2}  & \textbf{83.4} \\
    \hline
	\end{tabular}
    }
    \end{center}
    \label{tab:comparisons_with_sota_on_coco_test}
\end{table}

\subsection{Main Results}\label{sec:main_result}

\noindent\textbf{Gains on low-resolution backbone.}
In this part, we show the great improvement of \handle on non-HRNet paradigm backbone which encode the input image as low-resolution representation. All models and training settings are tightly aligned. The input resolution of all models is $256\times192$.

In~\cref{tab:com_hr_sb_tf}, \handle with ResNet-50 significantly outperforms SimpleBaseline, and \textit{it is even higher than HRNet-W32}, while the computational cost is much lower. Apart from that, \handle with the lightweight backbone MobileNet-V2 can achieve comparable performance with SimpleBaseline using ResNet-50 backbone. In contrast, the performance of the MobileNet-V2 based SimpleBaseline is much worse, $6.0$ AP lower than our method with the same backbone. It is worth noting that the computational cost of \handle with MobileNet-V2 is only about one-ninth that of SimpleBaseline with the same backbone.

\noindent\textbf{Comparison with the state-of-the-art methods.}
We compare the proposed \handle with state-of-the-art methods on COCO and MPII dataset.
\handle outperforms all regression-based and heatmap-based methods when using the same backbone, and achieves state-of-the-art performance with HRFormer-B backbone, \ie, $79.6$ AP on the COCO \emph{val} set and $78.3$ AP on the COCO \emph{test-dev} set.  
\handle with HRFormer-B can even outperform the previous state-of-the-art UDP-Pose ($384\times288$) by 1.1 AP on the COCO \emph{val} set, when using lower input resolution ($256\times192$). Quantitative results are reported in~\cref{tab:comparisons_with_sota_on_val} and \cref{tab:comparisons_with_sota_on_coco_test}.
On the MPII \emph{val} set, \handle with HRNet-W32 is 0.4 PCKh higher than heatmap-based method with HRNet-W32. Quantitative results are reported in~\cref{tab:mpii}.

\section{Conclusion}

We have proposed a novel pose estimation framework named \handle built upon Transformers, which largely improves the performance of the regression-based pose estimation and bypasses the drawbacks of heatmap-based methods such as the non-differentiable post-processing and quantization error.
Extensive experiments on the MS-COCO and MPII benchmarks show that \handle can achieve state-of-the-art performance among both regression-based methods and heatmap-based methods.

\clearpage
\bibliographystyle{splncs04}
\bibliography{egbib}



\section*{Supplementary material}

# Additional Results—Poseur: Direct Human Pose Regression with Transformers

Weian Mao[1]   Yongtao Ge[1,2]   Chunhua Shen[3]   Zhi Tian[1]   Xinlong Wang[1]
Zhibin Wang[2]   Anton van den Hengel[1]

[1] The University of Adelaide    [2] Alibaba Damo Academy    [3] Zhejiang University

## 1  The Effect of Training Schedules

In this section, we conduct experiments to show the effect of training schedules on the Poseur's performance, as shown in Tab. 1. In our paper, we use a longer training schedule (i.e., 325 epochs in total) than other methods, e.g., RLE [2] (270 epochs in total). In Tab. 1, we show that Poseur trained by 275 epochs or 250 epochs can also achieve impressive performance, which is only slightly lower than the fully-trained one in our paper. Thus, a longer training schedule is not the main reason for our superior performance.

| Epoch | $\text{AP}^{kp}$ | $\text{AP}^{kp}_{50}$ | $\text{AP}^{kp}_{75}$ | $\text{AP}^{kp}_M$ | $\text{AP}^{kp}_L$ |
|---|---|---|---|---|---|
| 150 | 74.1 | 90.1 | 81.3 | 67.4 | 76.8 |
| 175 | 74.6 | 90.2 | 81.7 | 67.9 | 77.2 |
| 200 | 74.8 | 90.3 | 81.7 | 68.0 | 77.6 |
| 225 | 75.0 | 90.3 | 81.8 | 68.2 | 77.8 |
| 250 | 75.2 | 90.7 | 82.3 | 68.4 | 78.0 |
| 275 | 75.3 | 90.3 | 82.3 | 68.5 | 78.2 |
| 300 | 75.4 | 90.4 | 82.6 | 68.6 | 78.4 |
| 325 | 75.5 | 90.7 | 82.7 | 68.7 | 78.3 |

Table 1: The effect of training schedules on the COCO *val* set

## 2  The Effect of Self-attention

In this section, we perform experiments to explore the effect of the self-attention module in the Poseur decoder. As shown in Tab. 2, the performance drops significantly from 75.5 AP to 74.0 AP when the self-attention module is removed from the decoder. Thus, we conjecture that the self-attention module can effectively model the relationship between different keypoints, improving Poseur's performance.

Moreover, We also visualize the self-attention weights across queries in Fig. 2. The left shoulder query attends to the most relevant keypoints, including left elbow, left wrist and left ear.

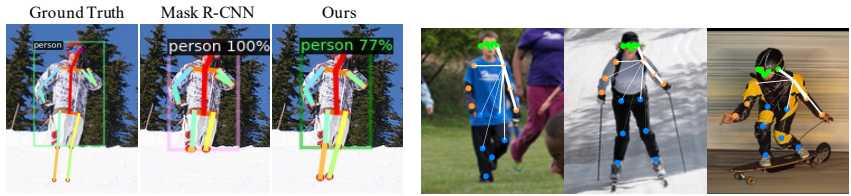

Fig. 1: Qualitative comparison on truncations. Heatmap-based methods (e.g.Mask R-CNN) can only predict keypoints within the bounding box, while Poseur can predict keypoints outside the bounding box

Fig. 2: Visualization of the self-attention weights between keypoint queries for left shoulder. Dots represent the keypoints. Lines depict attention weights between different joints. Thicker line indicates larger attention weight

| Self-Attn. | $AP^{kp}$ | $AP^{kp}_{50}$ | $AP^{kp}_{75}$ | $AP^{kp}_{M}$ | $AP^{kp}_{L}$ |
|---|---|---|---|---|---|
|  | 74.0 | 90.2 | 80.9 | 66.8 | 77.2 |
| ✓ | 75.5 | 90.7 | 82.7 | 68.7 | 78.3 |

Table 2: The effect of self-attention module on the COCO *val* set

| Share weight | Param. | $AP^{kp}$ | $AP^{kp}_{50}$ | $AP^{kp}_{75}$ | $AP^{kp}_{M}$ | $AP^{kp}_{L}$ |
|---|---|---|---|---|---|---|
|  | 32.3M | 75.5 | 90.7 | 82.7 | 68.7 | 78.3 |
| ✓ | 26.2M | 75.0 | 90.3 | 81.9 | 68.0 | 77.7 |

Table 3: The parameter reduction technique on the COCO *val* set

## 3    Reducing the Number of Parameters

Former works, e.g. Deeppose [5] and RLE [3], use fully-connected layers as decoder to regress keypoints, while Poseur has a transformer-based decoder. As the number of decoder layers increases, the model parameters increases rapidly, which may limit the deployment of Poseur for real-time applications that run on mobile devices.

In this section, we explore reducing the parameters of Poseur by sharing weights between different decoder layers. As shown in Tab. 3, the number of parameters of Poseur is significantly reduced, while the performance of Poseur only drops by 0.5 AP. Notably, the number of parameters of the backbone (ResNet-50) is 23.5 M, which means Poseur with weight sharing only introduces 2.7 M parameters.

## 4    Computational Cost of EMSDA

Let us denote the number of queries by $K$, and denote the number of pixels in the input feature maps $\{\mathbf{x}^l\}_{l=1}^{L}$ as $P$, and other notations follow our paper. The

| type | GFLOPs (Dec.) | $AP^{kp}$ | $AP^{kp}_{50}$ | $AP^{kp}_{75}$ | $AP^{kp}_M$ | $AP^{kp}_L$ |
|---|---|---|---|---|---|---|
| MSDA | 1.25 | 73.6 | 89.8 | 80.6 | 66.6 | 75.5 |
| EMSDA | 0.44 | 73.6 | 89.6 | 80.1 | 66.7 | 75.4 |

Table 4: Comparison between EMSDA and MSDA on the COCO *val* set. "GFLOPs (Dec.)": computatuional cost of the decoder

| Method | backbone | GFLOPs | FPS | Mem. Consumption | $AP^{kp}$ |
|---|---|---|---|---|---|
| RLE | HRNet-w32 | 7.1 | 62 | 1456M | 74.3 |
| Poseur | R-50 | **4.6** | **94** | **1386M** | **75.4** |

Table 5: Comparison between RLE and Poseur on the COCO *val* set. "Mem. Consumption": memory consumption of one image during the training stage

complexity of MSDA can be written as $O(KC^2 + PC^2 + 5KSC)$. Since $P$ is much larger than $K$, $C$ and $S$ (i.e., $P = 4080$ when the input image resolution is $256 \times 192$ and the feature maps from Res2 to Res5 are taken as the input), the computational cost mostly comes from the factor $O(PC^2)$. In our design, the EMSDA module significantly reduces the complexity to $O(KC^2+KC^2+5KSC)$, where $K \ll P$ (17 *vs.* 4080). As shown in Tab. 4, the performance of EMSDA is almost the same with that of MSDA, while EMSDA significantly reduces the computational cost from 1.25 GFLOPs to 0.44 GFLOPs.

## 5   Comparing the Performance of Poseur and RLE

As shown in Tab. 5, Poseur with ResNet-50 backbone achieves higher performance than RLE with HRNet-w32 backbone (75.4 AP *vs.* 74.3 AP), and has a faster inference speed than RLE (94 FPS *vs.* 62 FPS). The memory consumption of Poseur during the training is lower then that of RLE (1386 M *vs.* 1456 M). Although the memory consumption of Poseur during the testing is sightly higher than that of RLE (86.25 M *vs.* 68.12 M), the memory consumption of the whole system during the test (human detector and pose estimator) is exactly the same ($\sim 2000$ M) for most of methods in Tab.10 of the paper, including both Poseur and RLE.

## 6   Verifying the Effect of Keypoint Encoder and Query Decoder in Poseur

Compared to RLE [3], the proposed keypoint encoder and query decoder (without uncertainty estimation) can boost the performance by 3.8 AP on COCO [4]. This ablation study is performed with ResNet-50 [1]; all the settings are strictly aligned.

## 7   The Explanation of the Positional Encoding in Keypoint Encoder

Positional encoding in the proposed keypoint encoder transforms the coarse proposal $\hat{\boldsymbol{\mu}}_f \in \mathbb{R}^{K \times 2}$ from the x-y coordinates to the sine-cosine positional embedding. Denote an element in $\hat{\boldsymbol{\mu}}_f$ as $pos$, which is normalized to $[0, 2\pi]$, The positional encoding function can be written as $PE(pos, 2i) = sin(pos/10000^{2i/d})$; $PE(pos, 2i+1) = cos(pos/10000^{2i/d})$, where $d = 128$, $2i$ and $2i+1$ are the $2i^{th}$ and $(2i+1)^{th}$ dimension. In this way, a pair of x-y coordinates is transferred to two positional embeddings representing x and y axis respectively, which are concatenated to be the final encodings $\hat{\boldsymbol{\mu}}_f^* \in \mathbb{R}^{K \times 256}$.

## 8   Robustness to Truncation

Truncation is very common in real world scenes. We conduct qualitative visualization to show the superiority of our method. As depicted in Fig. 1, heatmap-based Mask R-CNN can only detect the joints inside the predicted boxes, while our method can infer the joints outside the boxes since the queries can attend to the whole input image.



\end{document}